\documentclass[sigconf]{acmart}   
\usepackage{multirow}
\AtBeginDocument{%
  }

\copyrightyear{2025} 
\acmYear{2025} 
\setcopyright{cc}
\setcctype{by}
\acmConference[MM '25]{Proceedings of the 33rd ACM International Conference on Multimedia}{October 27--31, 2025}{Dublin, Ireland}
\acmBooktitle{Proceedings of the 33rd ACM International Conference on Multimedia (MM '25), October 27--31, 2025, Dublin, Ireland}
\acmDOI{10.1145/3746027.3758309}
\acmISBN{979-8-4007-2035-2/2025/10}
\begin{document}

\title{Compressed Feature Quality Assessment: Dataset and Baselines}

\author{Changsheng Gao}
\email{changsheng.gao@ntu.edu.sg}
\affiliation{%
  \institution{Nanyang Technological University}
  \city{Singapore}
  \country{Singapore}
}
\author{Wei Zhou}
\email{ZhouW26@cardiff.ac.uk}
\affiliation{%
  \institution{Cardiff University}
  \city{Cardiff}
  \country{United Kingdom}
}
\author{Guosheng Lin}
\email{gslin@ntu.edu.sg}
\affiliation{%
  \institution{Nanyang Technological University}
  \city{Singapore}
  \country{Singapore}
}
\author{Weisi Lin\textsuperscript{\textdagger}}
\email{wslin@ntu.edu.sg}
\affiliation{%
  \institution{Nanyang Technological University}
  \city{Singapore}
  \country{Singapore}
}
\renewcommand{\shortauthors}{Changsheng Gao et al.}

\begin{abstract}
The widespread deployment of large models in resource-constrained environments has underscored the need for efficient transmission of intermediate feature representations. In this context, feature coding, which compresses features into compact bitstreams, becomes a critical component for scenarios involving feature transmission, storage, and reuse. However, this compression process inevitably introduces semantic degradation that is difficult to quantify with traditional metrics. To address this, we formalize the research problem of Compressed Feature Quality Assessment (CFQA), aiming to evaluate the semantic fidelity of compressed features. To advance CFQA research, we propose the first benchmark dataset, comprising 300 original features and 12000 compressed features derived from three vision tasks and four feature codecs. Task-specific performance degradation is provided as true semantic distortion for evaluating CFQA metrics. We systematically assess three widely used metrics -- MSE, cosine similarity, and Centered Kernel Alignment (CKA) -- in terms of their ability to capture semantic degradation. Our findings demonstrate the representativeness of the proposed dataset while underscoring the need for more sophisticated metrics capable of measuring semantic distortion in compressed features. This work advances the field by establishing a foundational benchmark and providing a critical resource for the community to explore CFQA. To foster further research, we release the dataset and all associated source code at \url{https://github.com/chansongoal/Compressed-Feature-Quality-Assessment}. 
\end{abstract}




\maketitle
\begin{figure*}[thp]
    \centering
    \includegraphics[width=0.95\linewidth]{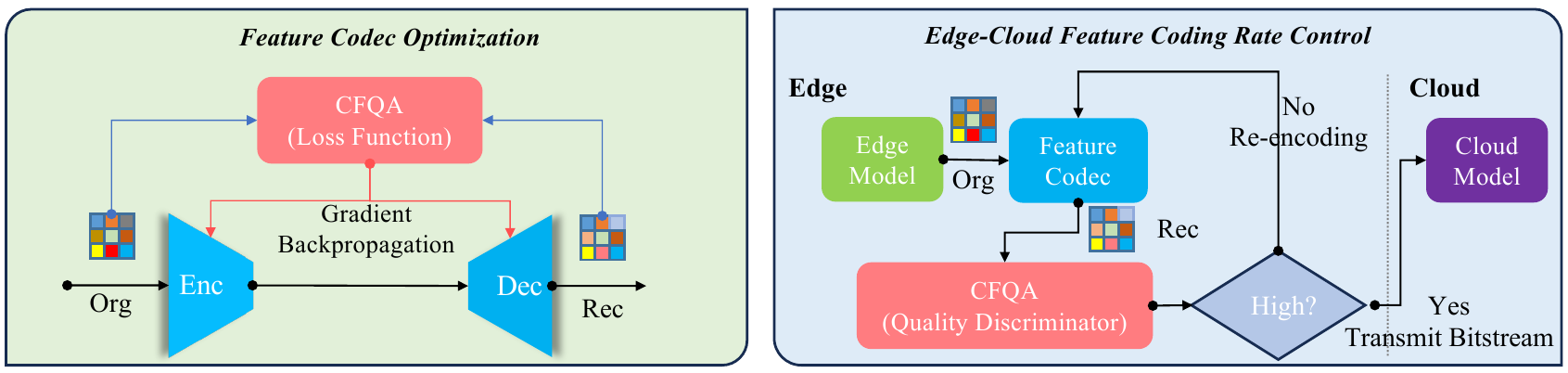}
    \caption{Exemplar application scenarios of compressed feature quality assessment.}
    \label{fig_application}
\end{figure*}
\section{Introduction}
Rapid deployment of large foundation models (e.g., DINOv2 \cite{oquab2023dinov2}, LLaMA3 \cite{dubey2024llama}) in distributed and resource-constrained environments has created a growing need to transmit intermediate features rather than raw signals \cite{gao2024feature}. Feature coding, which compresses these intermediate representations, plays a vital role in enabling scalable, privacy-preserving, and efficient systems. Unlike traditional image or video coding that prioritizes perceptual quality, feature coding targets the preservation of task-relevant semantics embedded in feature representations.

However, feature coding inevitably introduces semantic degradation: a loss of semantic information that may compromise the performance of large models. This degradation is fundamentally different from pixel-level distortions and often cannot be captured by conventional distortion metrics such as MSE or SSIM. Although task accuracy is a more reliable measurement of semantic distortion, it is impractical for practical deployments: downstream tasks may be inaccessible, costly to run, or unavailable in the feature coding process. These limitations highlight an urgent and underexplored problem: \textbf{Compressed Feature Quality Assessment (CFQA)} -- How can we estimate the semantic distortion of compressed features without relying on downstream inference?

Solving CFQA presents several challenges. First, there is no public dataset that provides compressed features with corresponding task performance across multiple tasks and codecs. Second, existing similarity metrics lack validation for high-dimensional features and often fail to generalize across various tasks. Third, the field lacks a unified evaluation protocol to assess whether a given metric reliably measures semantic distortion.

To bridge this gap, we take the first step towards a systematic study of CFQA. In contrast to previous works that rely on task supervision or task-specific codecs, we treat CFQA as a standalone problem and aim to benchmark its core components: dataset, metrics, and evaluation protocols.
In summary, we make the following contributions:
\begin{itemize}
    \item \textbf{Benchmark Dataset:} We construct the first comprehensive CFQA dataset, consisting of 300 original features and 12000 compressed features from three vision tasks (image classification (Cls), semantic segmentation (Seg), and depth estimation (Dpt)) and four representative codecs (handcrafted and learning-based). The dataset enables quantitative analysis of semantic distortion across tasks, bitrates, and coding methods.
    \item \textbf{Ground-truth Semantic Distortion:} For each compressed feature, we provide task-specific semantic distortion labels computed by comparing task head outputs using original and compressed features. These labels serve as the ground truth for training and evaluating semantic distortion metrics.
    \item \textbf{Baseline Metric Evaluation:} We evaluate three representative metrics (MSE, cosine similarity, and Centered Kernel Alignment (CKA)) and analyze their sensitivity to feature coding and their ability to measure task-specific semantic distortion. Our analysis reveals their varying strengths and limitations, providing insight into their applicability to semantic distortion measurement.
\end{itemize}
By establishing a standardized benchmark and evaluation framework, this work bridges the gap between low-level compression and high-level semantic utility. We hope to catalyze the development of lightweight, generalizable, and task-agnostic CFQA metrics that advance the real-world adoption of feature coding. Our dataset and code are publicly available to foster reproducibility and future research.

\section{Related Work}
\label{sec_related_work}
\subsection{Feature Coding}
Feature coding has received growing attention \cite{zhang2021MSFC,kim2023end,misra2022video,alvar2019multi,alvar2020bit,guo2023toward,fcm_cfp,yan2021SSSIC,choi2018deep,chen2020toward,chen2020data,gao2024dmofc,zhu2024learned,gao2024imofc,ma2024feature,chen2024end} in edge-cloud collaborative intelligence scenarios.
Recently, large model feature coding has attracted increasing interest \cite{gao2024feature,gao2025dtufc,gao2025cross}. These works extend feature coding to scenarios where features are transmitted, stored, and reused.  

\subsection{Semantic Distortion Measurement}
Semantic distortion measurement methods are classified into three main categories: signal fidelity metrics, semantics fidelity metrics, and task-based metrics.
Signal fidelity metrics \cite{choi2018deep, cai2022high, wang2022towards, niu2021entropy,kim2023end} focus on measuring the distortion between original and reconstructed features using traditional metrics like MSE. This approach is directly borrowed from image compression techniques, where the signal similarity is regarded as the compression distortion.

Semantics fidelity metrics \cite{alvar2021pareto,alvar2020bit,choi2021latent,ikusan2021rate,feng2022image,chen2019lossy,hu2020sensitivity,chen2020data,duan2020video} assess the preservation of semantic information, where distortion is measured based on the performance drop in specific machine vision tasks. These methods provide a more task-relevant measure of distortion, linking reconstruction quality directly to task performance.

Task-based metrics \cite{singh2020end,yang2021video,gao2024rethinking,zhang2021MSFC,yan2021SSSIC,alvar2019multi,misra2022video,henzel2022efficient,wang2023low,li2023attention,yuan2022feature} directly measure semantic distortion using the task performance. These methods are specialized for a particular task and lack generalizability.

\begin{table*}[]
    \centering
    \resizebox{\textwidth}{!}{%
     \begin{tabular}{@{}cccccccc@{}}
\toprule
\textbf{Task} & \textbf{Source} & \textbf{Num. of Org. Feat.} & \textbf{Pre-processing} & \textbf{Feature Codecs}                                                                                                                     & \textbf{Num. of Comp. Feat.} & \textbf{Feature Shape}                     & \textbf{GT Distortion} \\ \midrule
\textbf{Cls}  & ImageNet        & 100                       & Resize                & \multirow{3}{*}{\begin{tabular}[c]{@{}c@{}}HM, VTM\\ Multi-task Hyperprior\\ Task-specific Hyperprior\end{tabular}} & 4000                        & 257$\times$1536                    & Rank                   \\
\textbf{Seg}  & VOC 2012        & 100                       & Flip and Crop         &                                                                                                                                     & 4000                        & 2$\times$1370$\times$1536          & mIoU difference        \\
\textbf{Dpt}  & NYUv2           & 100                       & Flip                  &                                                                                                                                     & 4000                        & 2$\times$4$\times$1611$\times$1536 & RMSE difference        \\ \bottomrule
\end{tabular}
}
    \caption{Abstract information of the proposed dataset for CFQA. (Refer to Sec. \ref{sec_dataset} for more details.)}
    \label{tab_dataset}
\end{table*}

\section{Problem Formulation and Applications}
\subsection{Problem Formulation}
\label{subsec_problem}
Given a pretrained model \( \mathcal{M} \), we denote its extracted intermediate feature from an input \( x \in \mathcal{X} \) as \( \mathbf{f} = \mathcal{M}(x) \in \mathbb{R}^d \), where \( d \) is the feature dimension. A feature codec \( \mathcal{C} \) encodes \( \mathbf{f} \) into a compact bitstream and decodes it back to \( \hat{\mathbf{f}} = \mathcal{C}^{-1}(\mathcal{C}(\mathbf{f})) \). Our objective is to evaluate how much \textit{semantic information} is preserved in \( \hat{\mathbf{f}} \) with respect to \( \mathbf{f} \), \textit{without accessing downstream task ground-truth or executing inference}.

We define this evaluation task as Compressed Feature Quality Assessment:
    Given a pair of original and compressed features \( (\mathbf{f}, \hat{\mathbf{f}}) \), estimate the semantic quality \( q \in \mathbb{R} \) of the compressed feature, such that \( q \) strongly correlates with its performance on downstream tasks.

Assume a downstream task \( \mathcal{T} \) with a head network \( h_{\mathcal{T}} \), producing an output \( y = h_{\mathcal{T}}(\cdot) \). Let the task-specific performance metric be denoted as \( \mathcal{A}_{\mathcal{T}}(\cdot) \). When using compressed features, the task performance becomes: $ s = \mathcal{A}_{\mathcal{T}}(h_{\mathcal{T}}(\hat{\mathbf{f}}))$. 
This value \( s \in \mathbb{R} \) reflects the \textit{ground-truth semantic utility} of the compressed feature. However, computing \( s \) requires task execution and labels, which may be unavailable or expensive. The goal of CFQA is to estimate a score $q = Q(\mathbf{f}, \hat{\mathbf{f}})$, where \( Q(\cdot) \) is a \textit{task-agnostic quality metric}, such that \( q \approx s \) in terms of correlation across samples.

\subsection{Application Scenarios}
\label{subsec_application}
We illustrate two typical application scenarios of CFQA in Figure \ref{fig_application}. 
In the first scenario (left of Figure \ref{fig_application}), CFQA can be directly integrated into the codec training process as a supervisory signal to align compression objectives with downstream task performance. Recent studies~\cite{singh2020end,gao2024dmofc} show that task-aware feature coding benefits significantly from semantic-aware supervision. CFQA offers such guidance without requiring end-to-end task labels. As shown on the left of Figure~\ref{fig_application}, both the original and reconstructed features are input to the CFQA module, which estimates the semantic distortion. The distortion score is then used to generate gradients for optimizing the encoder and decoder. With a well-designed CFQA metric, the resulting codec learns to preserve semantic information even under low-bitrate constraints.

In the second scenario (right of Figure \ref{fig_application}), CFQA is essential in edge-cloud collaborative systems, where features are extracted at the edge and transmitted to the cloud for inference. Since the downstream model resides on the cloud side, semantic distortion cannot be directly measured at the edge. Here, CFQA is used as a proxy to estimate task-relevant semantic distortion. As illustrated on the right of Figure~\ref{fig_application}, the compressed feature is first evaluated by the CFQA module. If it is judged to be of high quality, the bitstream is transmitted to the cloud. Otherwise, the edge device re-encodes the feature at a higher bitrate to better preserve semantic information. This strategy ensures the reliability of transmitted features while reducing unnecessary bandwidth consumption.

Although our work focuses on semantic distortion due to compression, we emphasize that the value of CFQA extends beyond codec benchmarking: it serves as a crucial component in a variety of systems where features are extracted, compressed, transmitted, cached, or reused.

\section{Dataset Construction}
\label{sec_dataset}
\subsection{Overview}
Our proposed dataset is designed to support the analysis of \textit{semantic distortion} introduced by lossy compression. 
It includes three tasks, 300 original features, and 12000 compressed features from 4 feature codecs. The features cover diverse image processing methods and multiple feature extraction strategies. The overall information of the dataset is presented in Table \ref{tab_dataset}.

\subsection{Model and Task Selection}
Since current feature coding research primarily focuses on visual signals, we initiate the study of CFQA with visual features. We adopt DINOv2~\cite{oquab2023dinov2} as our backbone feature extractor due to its strong generalization capability and widespread adoption in general-purpose vision tasks. 
To cover a broad range of semantic distortion, we select three widely studied vision tasks: image classification (Cls), semantic segmentation (Seg), and depth estimation (Dpt). These tasks span coarse-to-fine semantic understanding: Cls focuses on image-level category prediction, Seg introduces spatial semantics with class-wise alignment at the pixel level, and Dpt requires detailed geometry prediction.

This multi-task approach is critical for evaluating the generalizability and sensitivity of CFQA metrics, ensuring they remain robust across different semantic requirements.

\subsection{Source Data Collection}
To ensure semantic diversity, we select 100 representative samples for each task. For Cls, we sample 100 ImageNet~\cite{deng2009imagenet} images from 100 distinct categories, each correctly predicted by the DINOv2 classifier. For Seg, we sample 100 images from the Pascal VOC 2012~\cite{everingham2010pascal} validation set, covering all 20 semantic categories. For Dpt, we sample 100 images from the NYUv2~\cite{Silberman2012indoor} dataset, spanning all 16 scenes.
This sampling strategy balances feature diversity with manageable dataset scale, enabling rigorous yet tractable evaluation.

\subsection{Original Feature Collection}
For each image, we extract task-specific intermediate features from DINOv2's designated split points. These split points are chosen based on their semantic richness and alignment with common practice in split computing.
For Cls, we resize the image to $224 \times 224$ and extract features from the 40$^\text{th}$ Vision Transformer (ViT) block, which produces features in the shape of $257 \times 1536$ ( 256 patch tokens and 1 class token).
For Seg, we flip the original image horizontally and extract features from the same 40$^\text{th}$ ViT block, resulting in $2 \times 1370 \times 1536$ features.
For Dpt, we collect multi-scale features from the 10$^\text{th}$, 20$^\text{th}$, 30$^\text{th}$, and 40$^\text{th}$ ViT blocks. The original and its horizontally flipped images generate a stacked tensor of shape $2 \times 4 \times 1611 \times 1536$.

The inclusion of diverse image pre-processing techniques and varying split points emulates real-world input variability, enabling rigorous evaluation of CFQA metric generalizability.

\subsection{Compressed Feature Collection}
To simulate different types and strengths of semantic distortion, we compress original features through four codecs. All original features are flattened into 2D arrays before encoding.
\paragraph{Handcrafted Codecs} 
We select two handcrafted codecs: \textbf{HM Intra coding} (configured with \textit{encoder\_intra\_main\_rext.cfg}) and \textbf{VTM Intra coding} (configured with \textit{encoder\_intra\_vtm.cfg}). Before encoding, we first uniformly quantize the original feature values to [0, 1023].
For both codecs, we use the YUV 4:0:0 (monochrome) format for feature coding and set the quantization parameters to $\in \{2, 4, 6, 8,\ldots, 16, 18, 20\}$ to simulate various bitrates (measured by Bits Per Feature Point, \textbf{BPFP}) and distortion levels.

\paragraph{Learning-Based Codecs} 
We adopt the Hyperprior model~\cite{balle2018variational} as it represents a milestone in learning-based feature coding. 
 Since most learning-based feature coding methods build upon this architecture, it serves as an ideal testbed for studying the semantic distortion characteristics of learning-based methods.

To investigate how optimization strategies affect compressed feature quality, we implement two distinct variants.
\textbf{Multi-Task Hyperprior Codec}: The codec is trained on features extracted from the three tasks.
\textbf{Task-Specific Hyperprior Codec}: The codec is trained exclusively on features extracted from a single task.
All codecs are optimized following the training protocols established in~\cite{gao2024feature}. The rate-distortion trade-off parameters ($\lambda$) can be found in our released models.

\subsection{Semantic Distortion Collection}
To establish a rigorous benchmark for evaluating CFQA metrics, we require ground-truth measurements of semantic distortion. We define the true semantic distortion as the performance degradation in downstream tasks when using compressed features $\hat{\mathbf{f}}$ versus original features $\mathbf{f}$. 
For \textbf{Cls}, we measure the deviation in prediction confidence by computing the \textit{rank} in the softmax function generated from $\hat{\mathbf{f}}$. The original feature $\mathbf{f}$ achieves perfect ranking (rank=1), with higher ranks indicating more severe semantic degradation.
For \textbf{Seg}, we compute the mIoU difference between the segmentation masks predicted from $\hat{\mathbf{f}}$ and $\mathbf{f}$.
For \textbf{Dpt}, we compute the RMSE difference between the depth maps predicted from $\hat{\mathbf{f}}$ and $\mathbf{f}$.

These task-specific scores provide quantitative measures of semantic distortion, serving as the foundation for assessing CFQA metric performance.

\begin{table*}[]
    \centering
    \begin{tabular}{@{}cc|cc|cc|cc|cc|cc@{}}
    \toprule
    \multicolumn{6}{c|}{\textbf{HM}}                                                                        & \multicolumn{6}{c}{\textbf{VTM}}                                                                       \\ \midrule
    \multicolumn{2}{c|}{\textbf{Cls}} & \multicolumn{2}{c|}{\textbf{Seg}} & \multicolumn{2}{c|}{\textbf{Dpt}} & \multicolumn{2}{c|}{\textbf{Cls}} & \multicolumn{2}{c|}{\textbf{Seg}} & \multicolumn{2}{c}{\textbf{Dpt}} \\ \midrule
    \textbf{BPFP}   & \textbf{Acc.}  & \textbf{BPFP}  & \textbf{mIoU}   & \textbf{BPFP}  & \textbf{RMSE}   & \textbf{BPFP}   & \textbf{Acc.}  & \textbf{BPFP}  & \textbf{mIoU}   & \textbf{BPFP}  & \textbf{RMSE}   \\ \midrule
    \textbf{32}     & \textbf{100}   & \textbf{32}    & \textbf{83.39}  & \textbf{32}    & \textbf{0.37} & \textbf{32}     & \textbf{100}   & \textbf{32}    & \textbf{83.39}  & \textbf{32}    & \textbf{0.37} \\ \midrule
    0.006           & 0.00           & 0.002          & 4.59            & 0.19           & 1.73            & 0.006           & 1.00           & 0.004          & 22.56           & 0.23           & 1.59            \\
0.008           & 0.00           & 0.004          & 11.39           & 0.37           & 1.40            & 0.01            & 1.00           & 0.01           & 40.18           & 0.36           & 1.36            \\
0.012           & 2.00           & 0.009          & 32.22           & 0.64           & 1.27            & 0.02            & 10.00          & 0.02           & 47.59           & 0.52           & 1.25            \\
0.02            & 6.00           & 0.02           & 47.79           & 1.01           & 1.00            & 0.03            & 13.00          & 0.04           & 55.05           & 0.75           & 1.23            \\
0.04            & 18.00          & 0.05           & 59.63           & 1.41           & 0.81            & 0.06            & 26.00          & 0.09           & 65.24           & 1.02           & 1.06            \\
0.08            & 25.00          & 0.11           & 67.87           & 1.82           & 0.63            & 0.11            & 44.00          & 0.17           & 72.34           & 1.26           & 0.92            \\
0.15            & 55.00          & 0.21           & 74.54           & 2.16           & 0.55            & 0.23            & 81.00          & 0.31           & 76.81           & 1.52           & 0.83            \\
0.33            & 84.00          & 0.41           & 78.40           & 2.55           & 0.47            & 0.49            & 92.00          & 0.56           & 79.79           & 1.85           & 0.73            \\
0.67            & 94.00          & 0.73           & 80.49           & 2.91           & 0.43            & 0.86            & 97.00          & 0.91           & 81.33           & 2.23           & 0.59            \\
1.12            & 97.00          & 1.13           & 81.86           & 3.28           & 0.40            & 1.26            & 98.00          & 1.28           & 82.04           & 2.67           & 0.52            \\ \bottomrule
    \end{tabular}
    \caption{Rate-accuracy performance of handcrafted codecs, with the first row showing original features' performance.}
    \label{tab_h26x}
\end{table*}
\begin{table*}
    \centering
    \begin{tabular}{@{}cc|cc|cc|cc|cc|cc@{}}
    \toprule
    \multicolumn{6}{c|}{\textbf{Multi-Task Hyperprior}}                                                      & \multicolumn{6}{c}{\textbf{Task-Specific Hyperprior}}                                                      \\ \midrule
    \multicolumn{2}{c|}{\textbf{Cls}} & \multicolumn{2}{c|}{\textbf{Seg}} & \multicolumn{2}{c|}{\textbf{Dpt}} & \multicolumn{2}{c|}{\textbf{Cls}} & \multicolumn{2}{c|}{\textbf{Seg}} & \multicolumn{2}{c}{\textbf{Dpt}} \\ \midrule
    \textbf{BPFP}   & \textbf{Acc.}  & \textbf{BPFP}  & \textbf{mIoU}   & \textbf{BPFP}  & \textbf{RMSE}   & \textbf{BPFP}   & \textbf{Acc.}  & \textbf{BPFP}  & \textbf{mIoU}   & \textbf{BPFP}  & \textbf{RMSE}   \\ \midrule
    \textbf{32}     & \textbf{100}   & \textbf{32}    & \textbf{83.39}  & \textbf{32}    & \textbf{0.37} & \textbf{32}     & \textbf{100}   & \textbf{32}    & \textbf{83.39}  & \textbf{32}    & \textbf{0.37} \\ \midrule
    0.01            & 0.00           & 0.0001            & 2.65            & 0.0003           & 1.68            & 0.34            & 16.00          & 0.05           & 50.67           & 0.12           & 2.09            \\
    0.12            & 18.00          & 0.10           & 58.74           & 0.10           & 1.79            & 0.43            & 35.00          & 0.08           & 60.69           & 0.18           & 1.66            \\
    0.58            & 44.00          & 0.46           & 76.70           & 0.52           & 1.55            & 0.56            & 55.00          & 0.14           & 65.45           & 0.30           & 1.22            \\
    0.71            & 52.00          & 0.57           & 77.80           & 0.64           & 1.24            & 0.65            & 61.00          & 0.21           & 71.89           & 0.42           & 0.88            \\
    1.01            & 59.00          & 0.85           & 79.30           & 0.91           & 0.99            & 1.15            & 70.00          & 0.29           & 74.76           & 0.57           & 0.82            \\
    1.19            & 67.00          & 1.01           & 80.06           & 1.07           & 1.45            & 1.39            & 72.00          & 0.42           & 78.10           & 0.70           & 0.73            \\
    1.34            & 77.00          & 1.14           & 80.50           & 1.20           & 0.77            & 1.81            & 78.00          & 0.65           & 79.09           & 1.06           & 0.60            \\
    1.45            & 81.00          & 1.23           & 80.66           & 1.29           & 0.92            & 1.96            & 81.00          & 0.94           & 79.59           & 1.30           & 0.49            \\
    1.75            & 94.00          & 1.45           & 81.22           & 1.52           & 0.76            & 2.15            & 84.00          & 1.41           & 81.02           & 1.39           & 0.44            \\
    2.18            & 96.00          & 1.77           & 81.73           & 1.84           & 0.42            & 2.36            & 88.00          & 1.65           & 82.02           & 1.48           & 0.43            \\ \bottomrule
    \end{tabular}
    \caption{Rate-accuracy performance of learning-based codecs, with the first row showing original features' performance.}
    \label{tab_hyper}
\end{table*}
\begin{table*}[]
    \centering
    \begin{tabular}{@{}c|c|cc|cc|cc@{}}
\toprule
                                     &               & \multicolumn{2}{c|}{\textbf{MSE}} & \multicolumn{2}{c|}{\textbf{Cosine Similarity}} & \multicolumn{2}{c}{\textbf{CKA}} \\ \midrule
\textbf{Codec}                       & \textbf{Task} & \textbf{PLCC}  & \textbf{SROCC}  & \textbf{PLCC}         & \textbf{SROCC}         & \textbf{PLCC}  & \textbf{SROCC}  \\ \midrule
\multirow{3}{*}{\textbf{HM}}                                                                   & \textbf{Cls}  & 0.6641         & 0.9113          & -0.7368               & -0.9116                & -0.5788        & -0.9089         \\
  & \textbf{Seg}  & -0.6210        & -0.6729         & 0.6866                & 0.6729                 & 0.6133         & 0.6726          \\
  & \textbf{Dpt}  & 0.8601         & 0.9036          & -0.8799               & -0.9036                & -0.8829        & -0.9027         \\ \midrule
\multirow{3}{*}{\textbf{VTM}}                                                                  & \textbf{Cls}  & 0.6220         & 0.8939          & -0.7089               & -0.8939                & -0.5499        & -0.8932         \\
  & \textbf{Seg}  & -0.4784        & -0.5213         & 0.5131                & 0.5213                 & 0.4885         & 0.5213          \\
  & \textbf{Dpt}  & 0.8344         & 0.8750          & -0.8572               & -0.8747                & -0.8478        & -0.8754         \\ \midrule
\multirow{3}{*}{\textbf{\begin{tabular}[c]{@{}c@{}}Hyperprior\\ (Multi-Task)\end{tabular}}}   & \textbf{Cls}  & -0.0281        & 0.4707          & -0.8907               & -0.7165                & -0.3277        & -0.6593         \\
  & \textbf{Seg}  & 0.0595         & -0.0251         & 0.3496                & 0.0604                 & -0.0059        & 0.0847          \\
  & \textbf{Dpt}  & 0.6466         & 0.6675          & -0.4752               & -0.6528                & -0.5876        & -0.6004         \\ \midrule
\multirow{3}{*}{\textbf{\begin{tabular}[c]{@{}c@{}}Hyperprior\\ (Task-Specific)\end{tabular}}} & \textbf{Cls}  & -0.0220        & 0.1303          & -0.5486            & -0.6084                & -0.2622        & -0.3365         \\
  & \textbf{Seg}  & 0.3259	&0.3235	&-0.0552	&-0.1669	&-0.3138	&-0.2187          \\
  & \textbf{Dpt}  & -0.0787        & 0.4915          & -0.8427               & -0.8258                & -0.5483        & -0.5638    \\ \bottomrule    
\end{tabular}
    \caption{Average CFQA performance of the three baseline metrics on Cls, Seg, and Dpt tasks.}
    \label{tab_corr}
\end{table*}
\section{Experiments and Analysis}
\subsection{Baseline CFQA Metrics}
To cover diverse types of distortions and similarity relationships between features, we select three complementary metrics: MSE, cosine similarity, and CKA ~\cite{kornblith2019similarity}. These metrics collectively span from local element-wise to global structural comparisons.
\begin{itemize}
    \item \textbf{MSE}: MSE measures feature distortion at the element level. It is widely used in signal processing fields.
    \item \textbf{Cosine similarity}: Cosine similarity measures the angular difference between feature vectors, normalized by their magnitudes. This metric captures directional alignment in the feature space, which is often more robust to scale distortions and is considered to better reflect semantic similarity in high-dimensional representations~\cite{radford2021learning}. 
    \item \textbf{CKA}: CKA measures similarity between two features using normalized HSIC (Hilbert-Schmidt Independence Criterion). It captures higher-order statistical relationships, making it particularly suitable for comparing architectural differences and global feature patterns.
\end{itemize}

\begin{figure*}[h]
    \centering
    \setlength{\tabcolsep}{2pt}
    \resizebox{0.95\textwidth}{!}{%
    \begin{tabular}{cccc}
        \includegraphics[height=4cm]{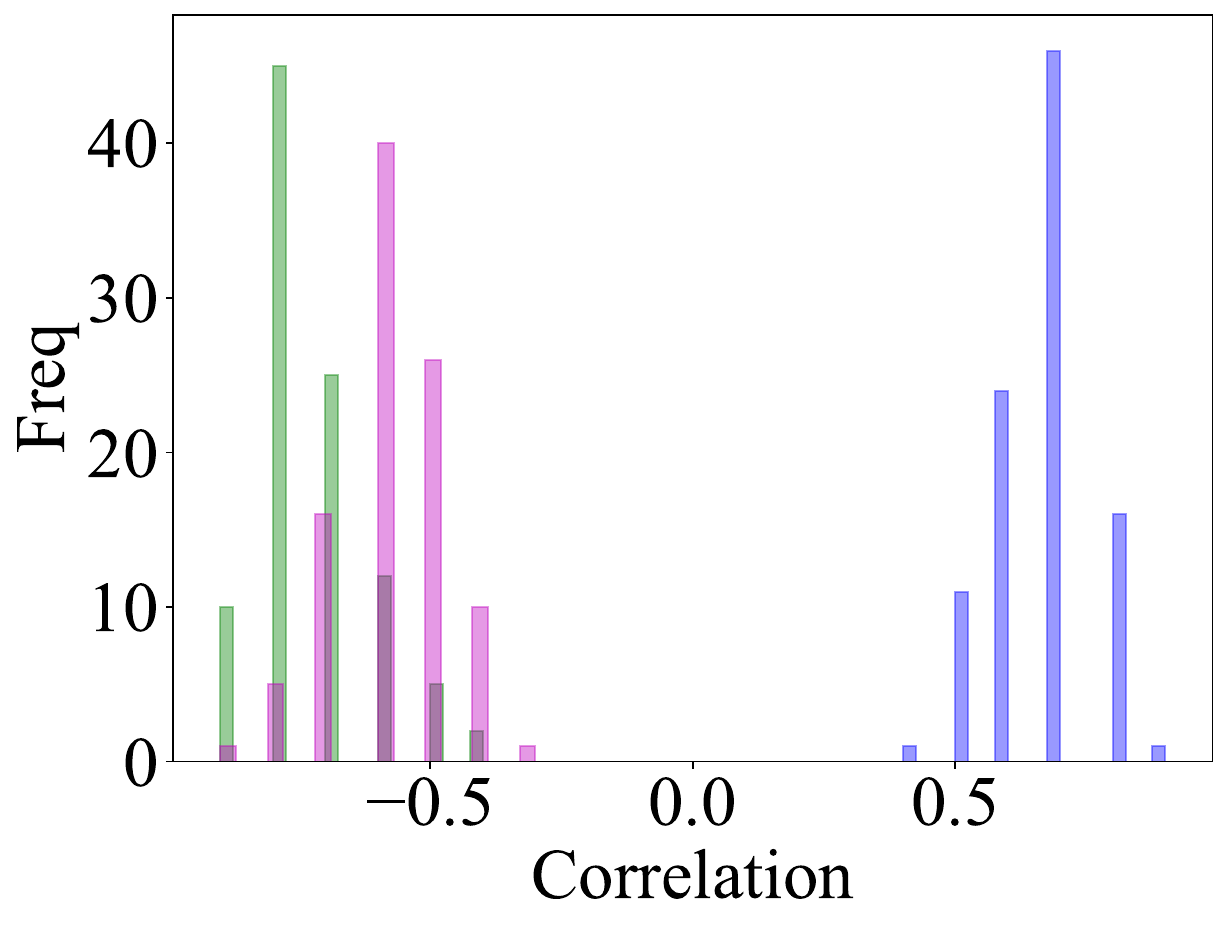} &
        \includegraphics[height=4cm]{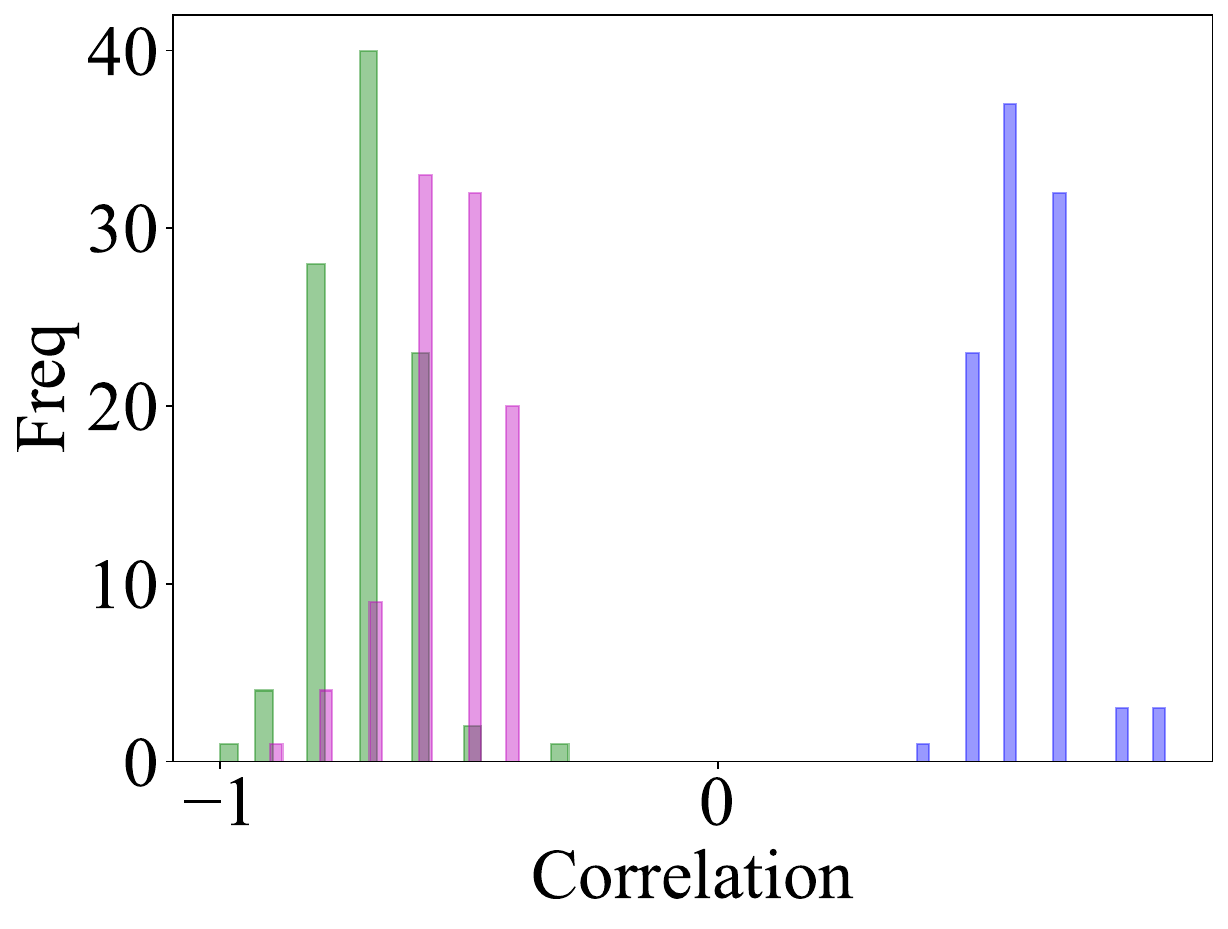} &
        \includegraphics[height=4cm]{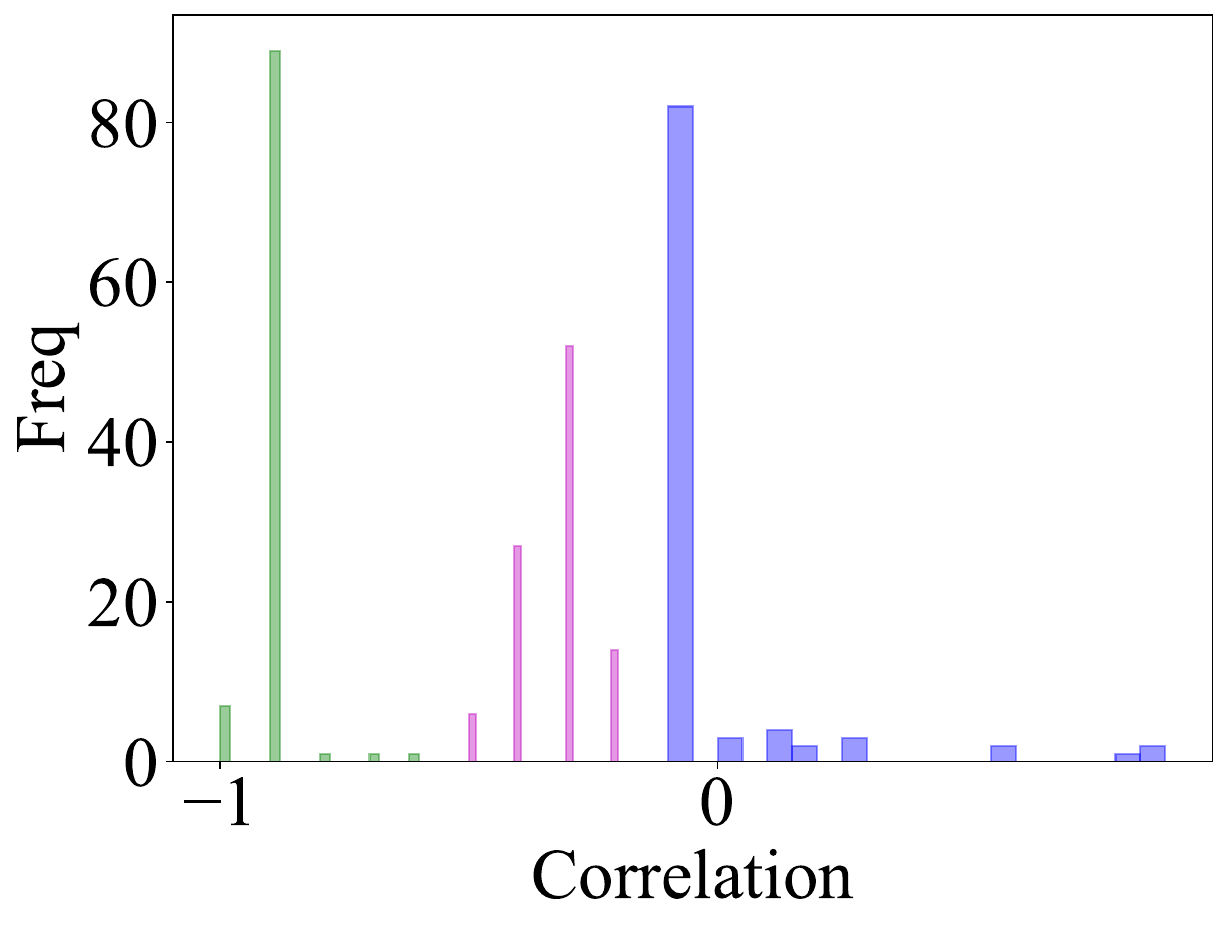} & 
        \includegraphics[height=4cm]{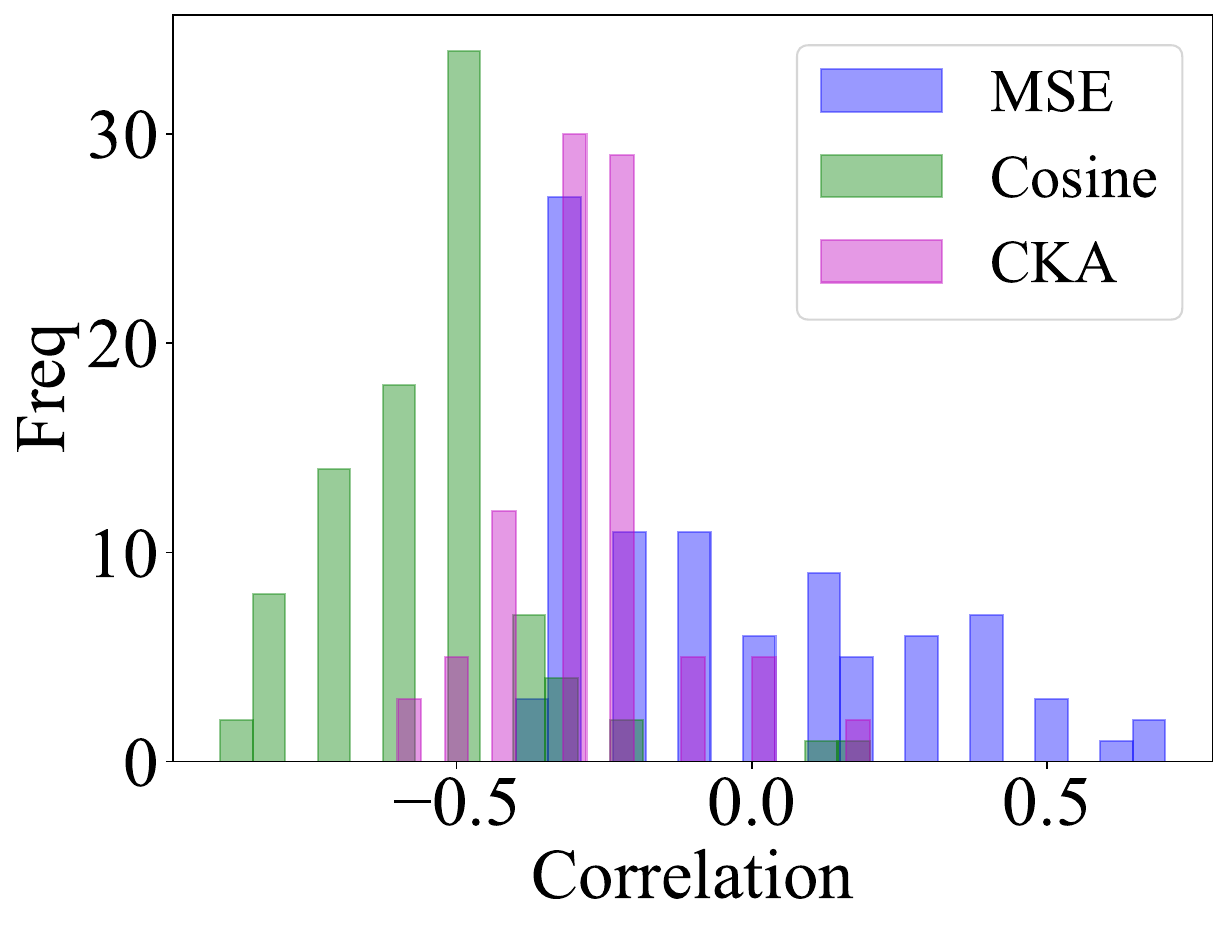} \\
        \includegraphics[height=4cm]{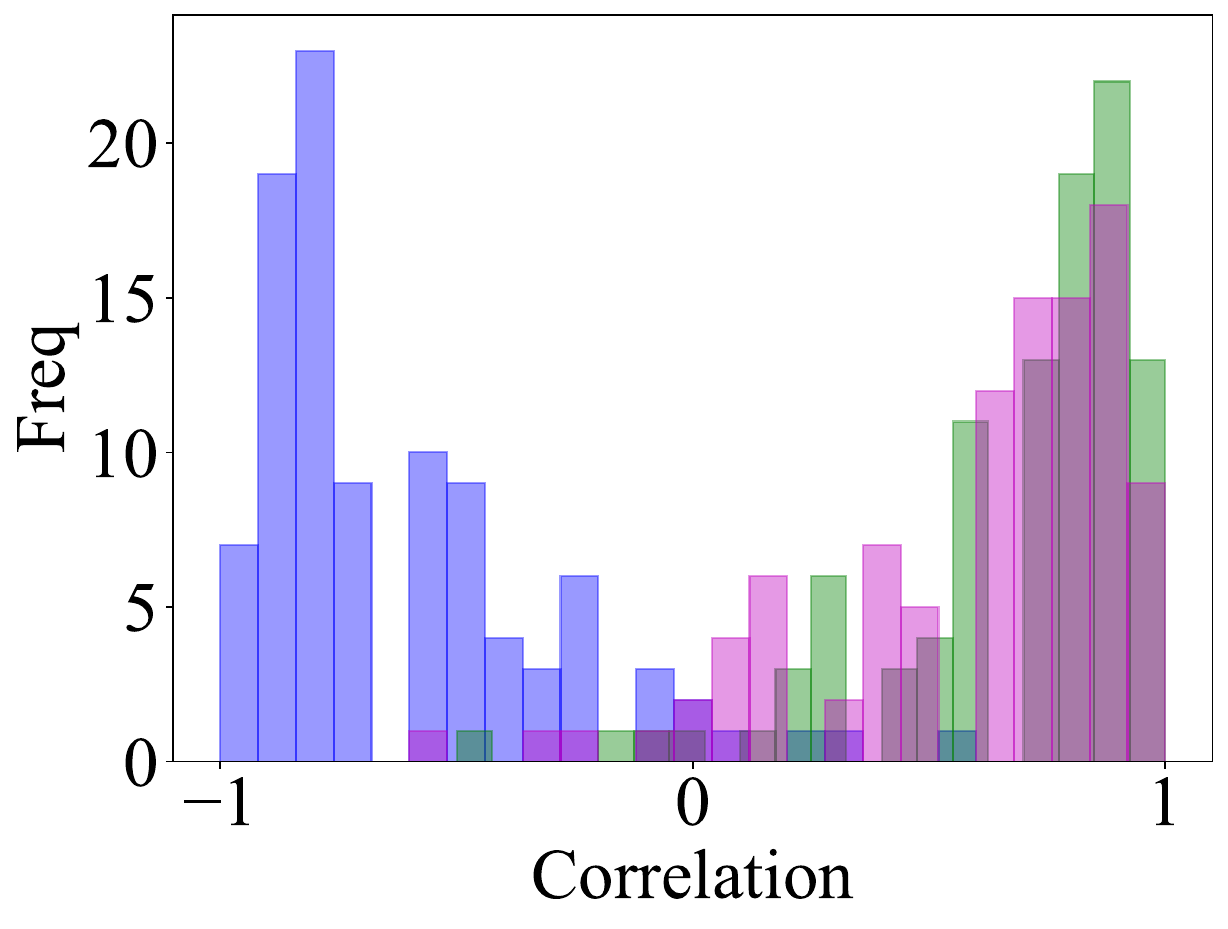} &
        \includegraphics[height=4cm]{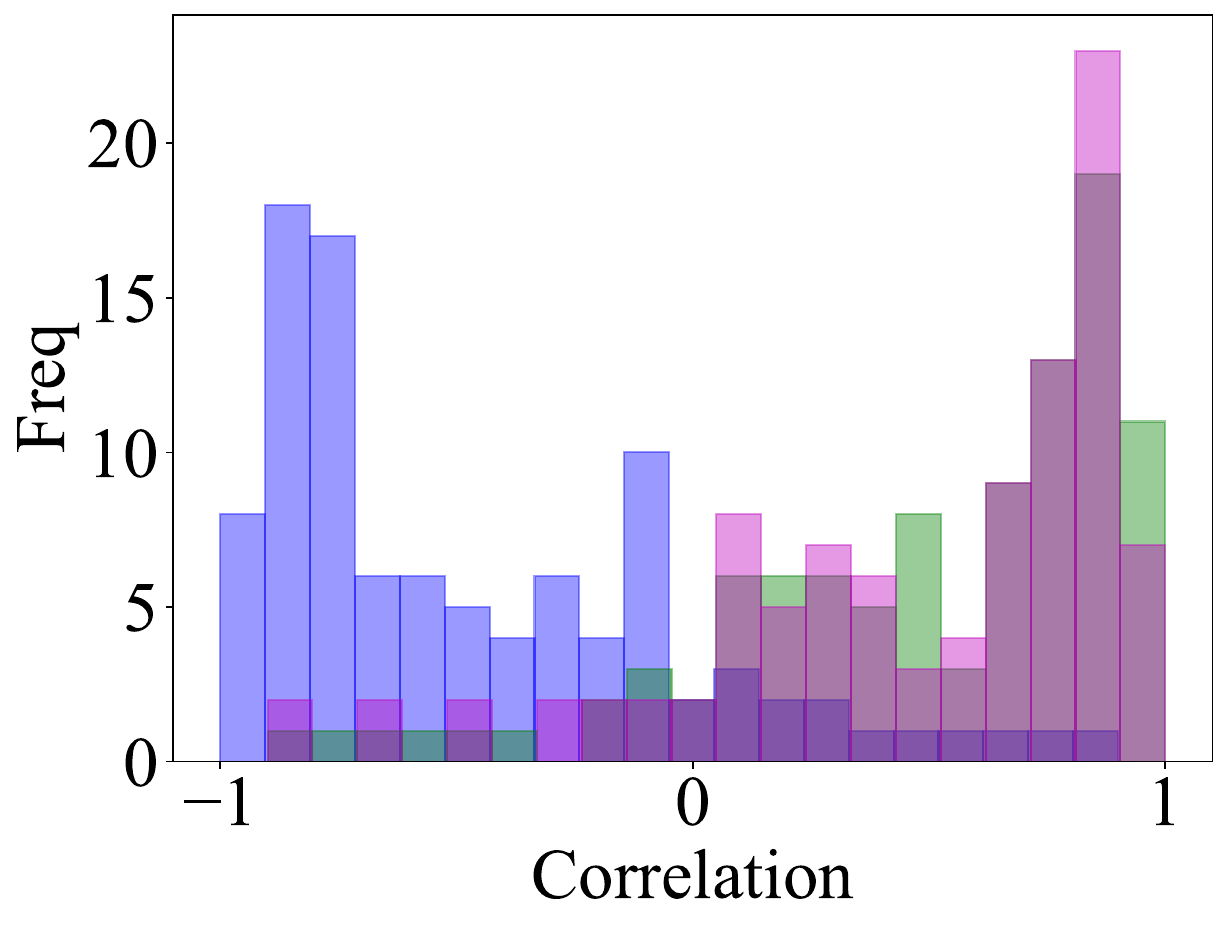} &
        \includegraphics[height=4cm]{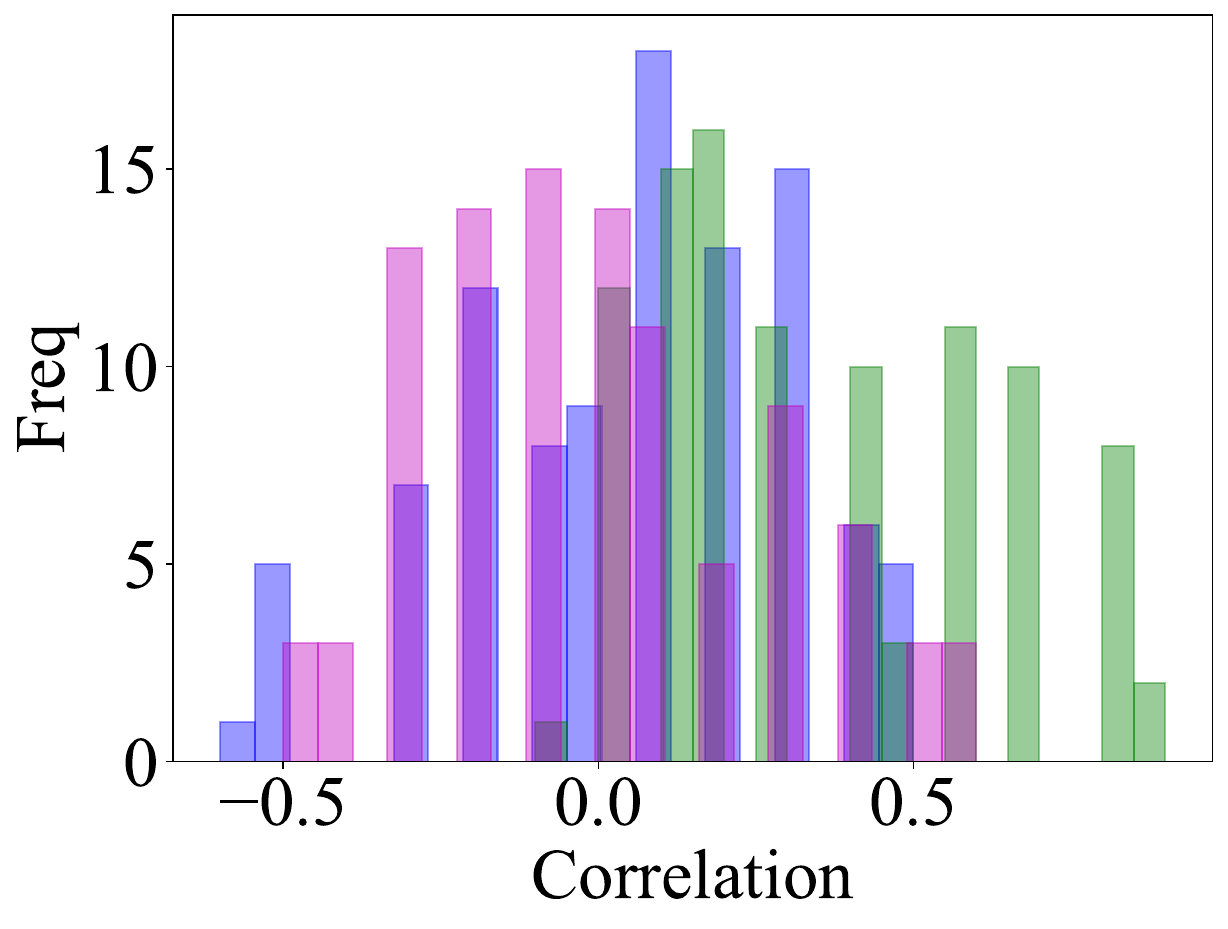} & 
        \includegraphics[height=4cm]{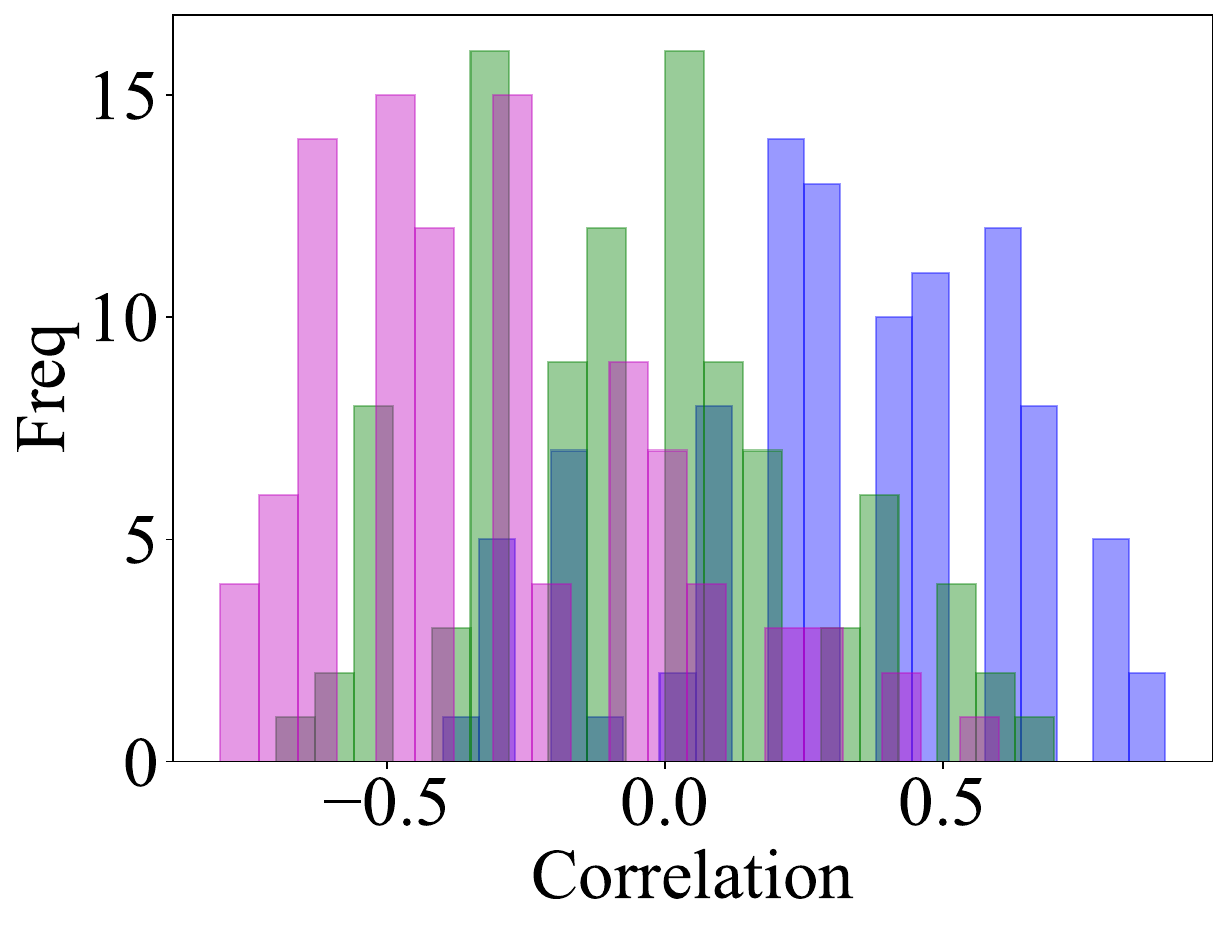} \\
        \includegraphics[height=4cm]{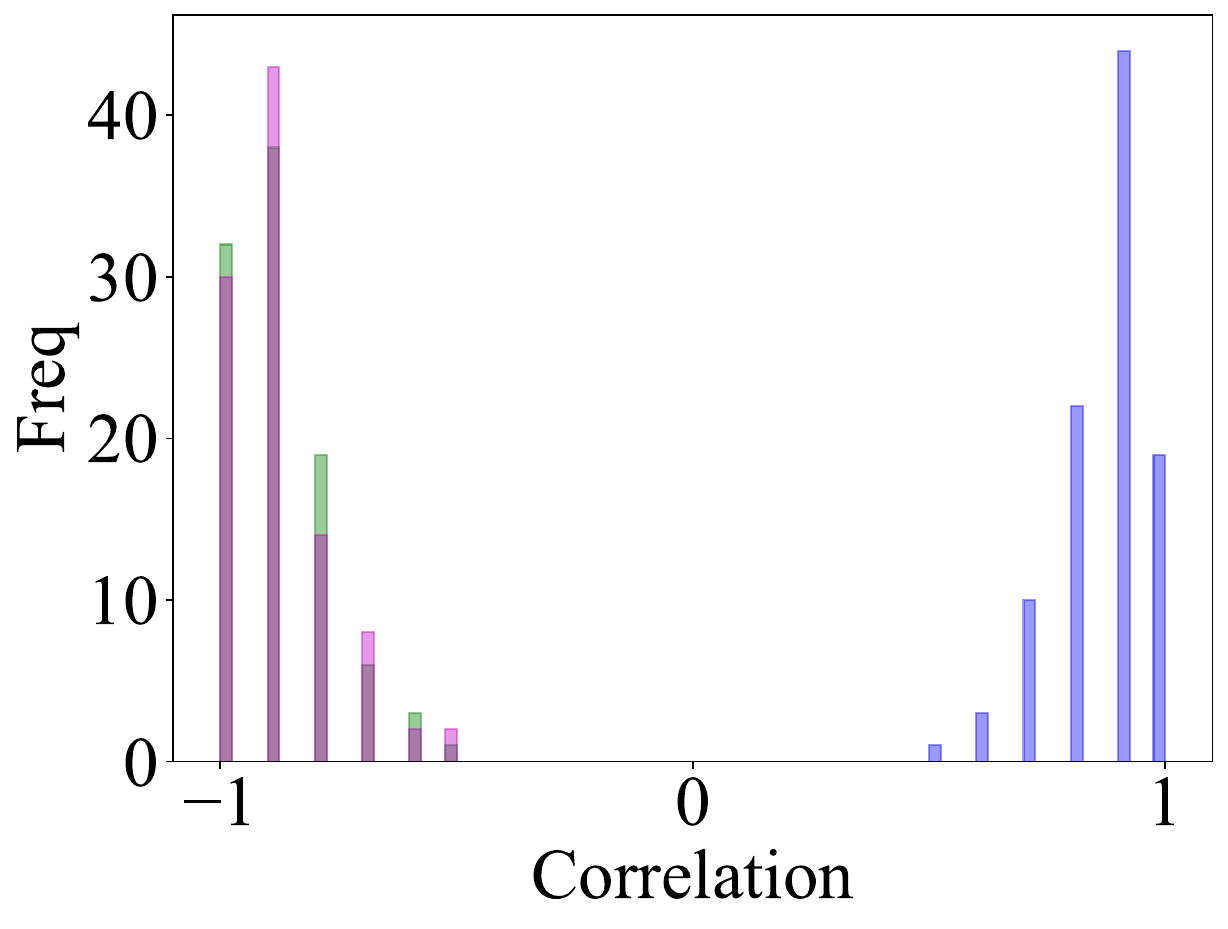} &
        \includegraphics[height=4cm]{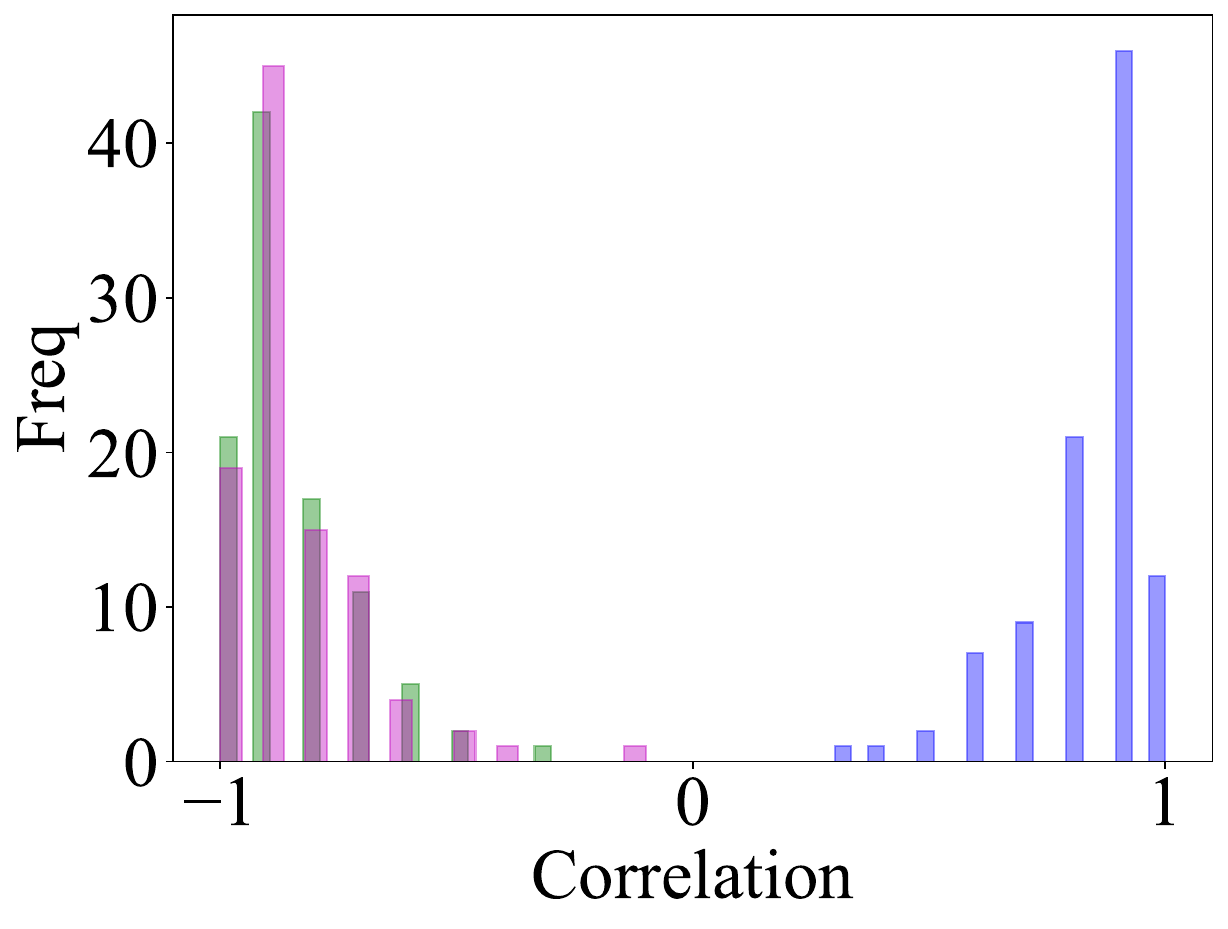} &
        \includegraphics[height=4cm]{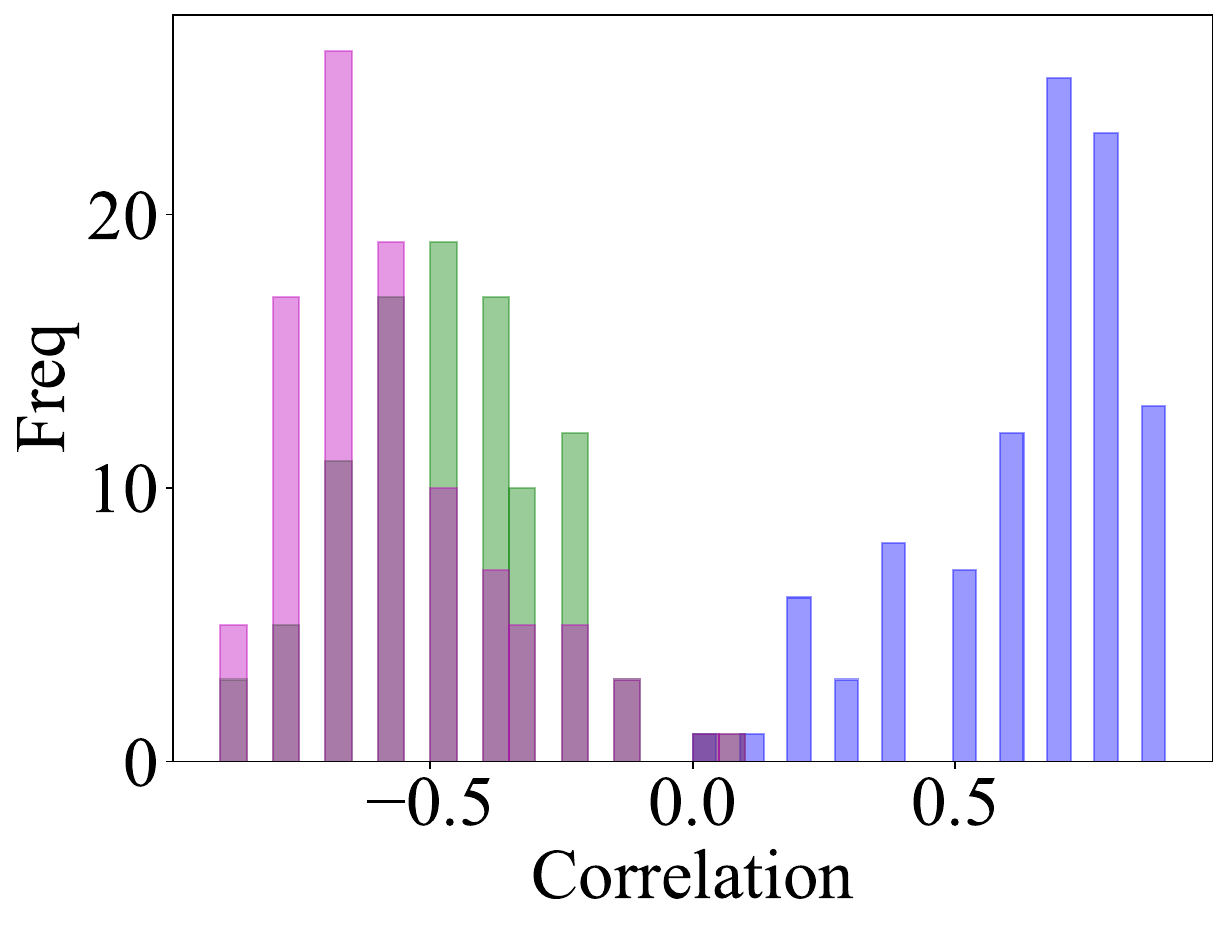} & 
        \includegraphics[height=4cm]{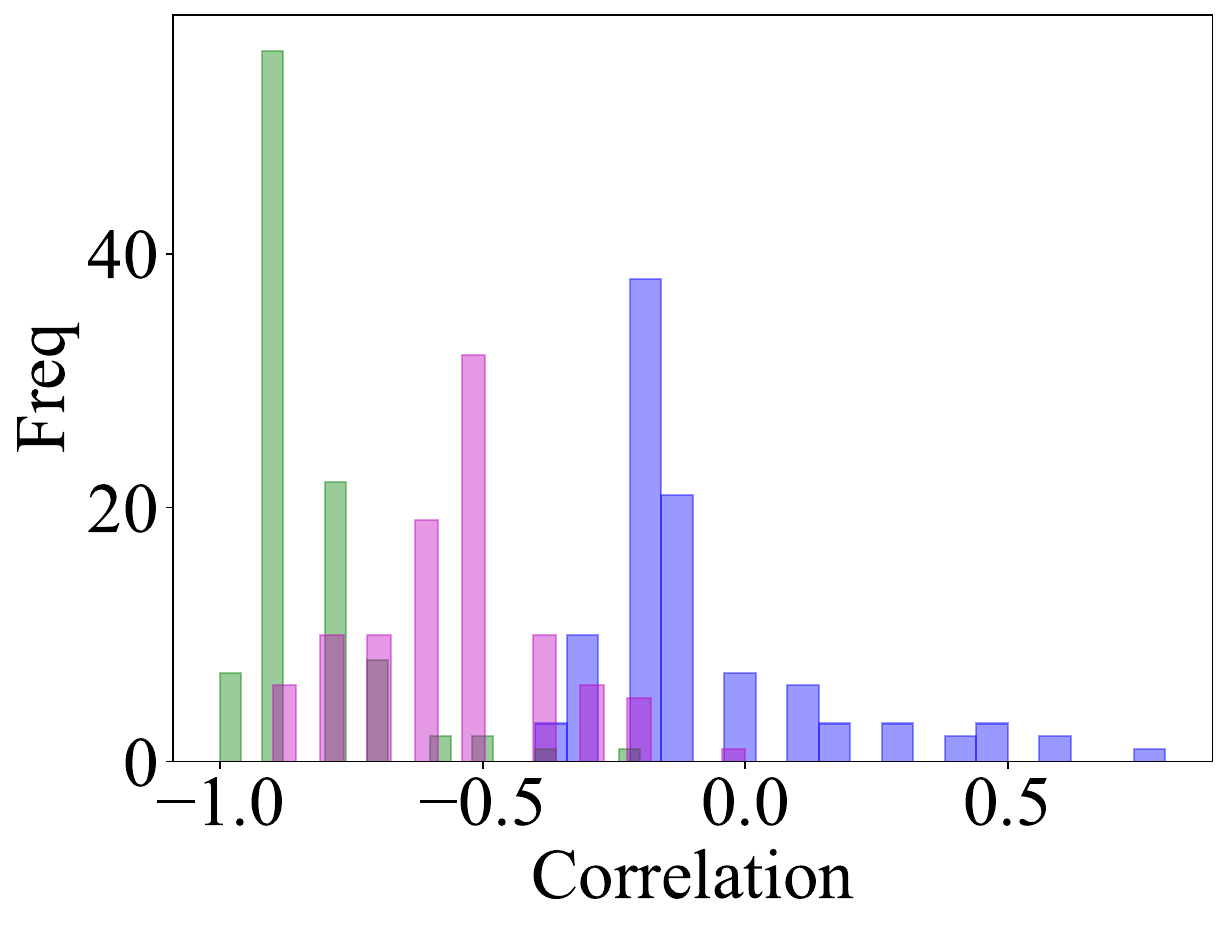} \\
    \end{tabular}
    }
    \caption{PLCC distribution visualization of MSE, cosine similarity, and CKA. The first, second, and third rows correspond to the Cls, Seg, and Dpt tasks, respectively. The first, second, third, and fourth columns correspond to the HM, VTM, multi-task-trained Hyperprior, and task-specific-trained Hyperprior codecs. }
    \label{fig_distribution}
\end{figure*}

\subsection{Evaluation Protocol}
\label{sec_protocol}
We define an evaluation protocol to assess how well a CFQA metric predicts the semantic distortion of compressed features. PLCC and SROCC are used in our experiments. 
\begin{itemize}
    \item \textbf{PLCC (Linearity)}: PLCC quantifies how well the predicted quality scores approximate the actual semantic distortion in a linear sense.
    \item \textbf{SROCC (Monotonicity)}: SROCC assesses the consistency in the ranking between predicted and true distortions.
\end{itemize}
Specifically, for each original feature $\mathbf{f}_i$, we compute these two metrics on its 10 predicted quality scores and 10 true semantic distortions.
Each metric is evaluated on all three tasks and across all codecs. 

\subsection{Rate-Accuracy Performance Analysis}
Table \ref{tab_h26x} presents the rate-accuracy performance of the handcrafted codecs. For all three tasks, both codecs exhibit a broad range of bitrates and corresponding accuracy levels, spanning from nearly lossless reconstruction to significant performance degradation. This wide variation highlights their effectiveness in simulating traditional coding distortions.
However, the Dpt task exhibits less variation in both bitrate and performance. This difference stems from the use of multi-scale features, which include lower-level layers with higher redundancy and reduced semantic abstraction. These characteristics lead to smaller fluctuations in both bitrate and distortion.

Table \ref{tab_hyper} shows the rate-accuracy performance of two learning-based codecs.  
Unlike the handcrafted codecs, the learning-based codecs show more diverse performance patterns across tasks. The multi-task Hyperprior codec achieves better performance than the task-specific Hyperprior codec for Cls, while the task-specific Hyperprior codec performs better in Dpt.
These results indicate that incorporating fine-grained task features improves the codec optimization of coarse-grained features, whereas fine-grained features benefit more from training on a single task-specific distribution.

These four codecs comprehensively simulate a wide range of compression distortions, making them not only highly suitable but also essential for supporting research in compressed feature quality assessment.

\subsection{CFQA Performance Analysis}
\subsubsection{Average Performance Analysis}
Table \ref{tab_corr} presents the average CFQA performance of the three baseline metrics. Overall, the three baseline metrics show higher PLCC and SROCC values for the handcrafted codecs compared to the learning-based codecs. This indicates that the semantic distortions introduced by handcrafted codecs are more stable. The larger fluctuations in semantic distortion observed for learning-based codecs are likely due to the inherent variability in the training process, as these models are data-driven and task-specific. Additionally, since handcrafted codecs are block-based, they tend to exhibit more consistent and predictable distortion patterns.

Among the three baseline metrics, cosine similarity consistently demonstrates a higher degree of linearity and monotonicity in relation to ground-truth semantic distortion. This is due to the metric's ability to capture angular relationships between feature vectors, making it particularly effective for high-dimensional features such as those produced by ViTs.

For handcrafted codecs, the three baseline metrics show better performance in capturing HM-generated distortions compared to VTM. This is likely because VTM employs more complex coding tools, resulting in more intricate and potentially less predictable distortion patterns. In contrast, for learning-based codecs, significant performance variation is observed across the three baseline metrics. This variability is expected, as learning-based encoders produce a wider range of distortion types, making it more challenging for simple metrics to achieve linear fitting.

Among the three tasks, Seg presents the most complex and difficult-to-fit distortions. For most codecs, the baseline metrics show poorer fitting for Seg compared to Cls and Dpt. This reflects the higher semantic complexity in Seg, which is more challenging for these metrics to capture.

\subsubsection{Distribution Analysis}
Figure~\ref{fig_distribution} visualizes the distribution of PLCC. To provide a clearer view, all PLCC values are rounded to the nearest tenth before computing the frequency histograms.

Overall, the PLCC distributions for handcrafted codecs are more concentrated compared to those of learning-based codecs. This aligns with the fact that handcrafted codecs introduce more consistent and predictable semantic distortion patterns. 

Among the three baseline metrics, cosine similarity exhibits the most concentrated PLCC distributions. This observation aligns with its superior average PLCC values, as reported in Table~\ref{tab_corr}, indicating that cosine similarity provides a more stable semantic distortion measurement across varying conditions. 

Across all three metrics, Seg shows more dispersed PLCC distributions compared to Cls and Dpt. In some cases, such as with CKA, both positive and negative correlations are observed within the same metric. This highlights the high complexity of distortion patterns in segmentation features. The broader distribution further confirms that existing baseline metrics struggle to model such complex semantic degradation accurately.

\subsection{Discussion}
Our evaluation reveals three key limitations of current CFQA metrics. First, signal-based metrics like MSE show limited task sensitivity and often correlate poorly with actual semantic degradation. Second, while cosine similarity and CKA capture structural relationships, they demonstrate instability when handling nonlinear distortions from learned codecs. Most importantly, none of these metrics achieves consistent performance across all tasks and compression methods, indicating that conventional similarity measures alone cannot provide universal CFQA solutions. These findings underscore the necessity for developing more adaptive, task-aware quality estimators -- a direction we plan to explore in future work.

\section{Conclusion}
This paper introduces the concept of Compressed Feature Quality Assessment, a crucial research area for evaluating the semantic distortion of compressed features in systems where features are transmitted, stored, and reused. We present the first benchmark dataset for CFQA, which lays the groundwork for further research in this field. 
We assess the widely used metrics in CFQA and provide insights for interested researchers.
Moving forward, we will focus on developing adaptive CFQA metrics capable of generalizing across diverse tasks and coding strategies.

\begin{acks}
This work was supported by the Ministry of Education of Singapore under Grant T2EP20123-0006. We acknowledge the support of GPU cluster built by MCC Lab of Information Science and Technology Institution, USTC.
\end{acks}


\bibliographystyle{ACM-Reference-Format}
\balance
\bibliography{refs}

\end{document}